# High-payload and self-adaptive robotic hand with 1-degree-of-freedom translation/rotation switching mechanism

Toshihiro Nishimura[1], *Member, IEEE*, Tsubasa Muryoe[2], and Tetsuyou Watanabe[1], *Member, IEEE*

*Abstract*—This study proposes a novel robotic hand that can achieve self-adaptive grasping and a large payload (over 20 kg) with a single actuator. Accordingly, two novel mechanisms, an actuation system with self-motion switching and a self-adaptive finger with a self-locking mechanism, are installed in a 1-degree-of-freedom robotic hand. The actuation system switches the output motion from translational to rotational according to the applied external load. The finger is bent by inserting a flexible shaft inside it. Its bending posture can conform to the shape of the object owing to the flexible shaft, and the posture is fixed by a self-locking mechanism, which can be released by the rotational motion of the actuation system. This study presents a mechanical analysis of these mechanisms to achieve the desired behavior. The analysis was validated experimentally, and a robotic hand with these mechanisms were evaluated using grasping tests.

*Index Terms*—Grasping, grippers and other end-effectors, mechanism design, grasping

## I. INTRODUCTION

THIS study presents a novel three-fingered robotic hand with self-adaptability and a large payload driven by a single actuator. A self-adaptive function is important so that robots can grasp objects of various shapes. Self-adaptability can be achieved using deformable elements, such as springs and wires, to construct robotic fingers and hands. However, they can only support a small payload because a large load can cause the finger elements to deform significantly, making it difficult to maintain the grasping posture. If a high-power motor is used and the stiffness of the finger elements increases, the payload increases, but the adaptability decreases, and the size of the robotic hand also increases. Hence, both a self-adaptive function and a large payload are difficult to achieve simultaneously. To address this issue, a self-locking mechanism resistant to external disturbances is employed in a self-adaptive robotic hand. Furthermore, it is also challenging to address this issue using a single low-power motor to minimize the overall system cost. The developed robotic hand is shown in Fig. 1. This robotic hand has two key mechanisms: a 1-degree-of-freedom (DOF) actuation mechanism that can switch the translation and rotation motions, and a self-adaptive finger with a self-locking mechanism (Fig. 1(a)). The translation/rotation switching mechanism (T/R-SW mechanism) switches its motion according to the external load applied to the tip of the drive shaft (purple part in Fig. 1(a)) in the mechanism. This mechanism generates a translational motion when the applied load is small. When the magnitude of the load exceeds the threshold value, the motion becomes rotational. The forward rotation of the motor translates the drive shaft of the T/R-SW mechanism. The drive shaft is connected to a flexible shaft (red and gray parts in Fig. 1(a)). This flexible shaft is installed inside the finger, where it acts as a spine, such that the finger bends with its translation. The flexibility of this shaft allows the fingers to bend in a self-adaptive manner upon contact with an object or environment (Fig. 1(b)). The finger also has a self-locking mechanism to prevent its extension. Therefore, the finger bends self-adaptively in the flexion direction, while the finger motion is locked in the extensional direction. The bent-finger posture is maintained even when disturbing forces are applied (Fig. 1(a)). Thus, a large payload is realized with this design. The release of the grasped object, i.e., the unlocking of the self-locking mechanism and the return of the finger to its initial posture, is accomplished by simply reversing the motor. In the T/R-SW mechanism, the reverse translational motion is prevented by the locking mechanism in the finger. If the motor torque is increased, even when locked, the motion of the drive shaft switches from translational to rotational, and the locking is released. After unlocking, the motion of the T/R-SW mechanism is switched from rotational to translational because the load that restricts the translational motion is removed. The T/R-SW mechanism enables the robotic hand to generate a grasping motion and operate the locking mechanism using a single actuator (Fig. 1(c)). The 1-DOF actuation system also realizes a lightweight robotic hand design and a simple control scheme.

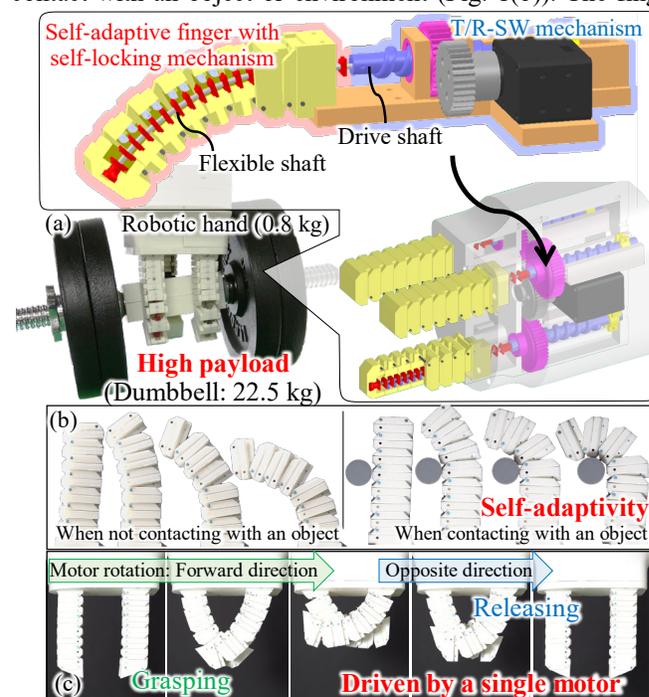

**Fig. 1.** Proposed robotic hand. (a) Design and key mechanisms. The payload of the proposed robotic hand is greater than 20 kg. (b) Self-adaptive function. (c) Behavior of the robotic hand using 1-DOF motor control.

### A. Related work

Deformable and flexible elements allow robotic hands to conform to the shape of an object and achieve a stable grasping posture. Several robotic hands with self-adaptive grasping functions have been developed. In [1] and [2], a linkage mechanism, including springs, was used to achieve self-adaptive grasping. Raymond et al. developed a tendon-driven robotic hand in which finger links were connected via compliant joints for self-adaptive grasping [3][4]. Robotic hands based on a fin-ray structure composed of deformable and flexible materials provide a self-adaptive grasping function [5][6]. Pneumatic soft robotic hands also exhibit self-adaptability [7][8][9]. Deformability caused by flexible elements and materials is effective for self-adaptive functions. However, it causes a reduction in the payload of the robotic hand due to the large deformation by the large payload.

Several studies have attempted to develop robotic hands with high payloads. Takayama et al. developed a robotic hand that generated a large grasping force using a force magnification mechanism [10]. Jeong et al. designed a robotic hand with a large grasping force by switching the grasping mode between the force and speed modes [11]. Our group also developed robotic hands with high payloads using a lock mechanism [12] and a mechanism inspired by a chuck device [13]. As in [12], locking mechanisms are well-known for providing large payloads to robotic hands. There are two types of locking mechanisms: active locking and self-locking. In active locking, actuators are required to lock and unlock the motion of the robotic hand [14][15]. In self-locking, locking is activated automatically according to robotic hand motion and unlocking is activated via actuators [12][16][17][18]. In both types, at least one actuator is required for locking or unlocking in addition to the actuators required to drive the fingers.

Although a locking mechanism was installed on a self-adaptive robotic hand to achieve a high payload in [15][16][17][18], no attempt has been made to realize a robotic hand with both self-adaptive grasping and a high payload using a single actuator. Therefore, a self-adaptive finger with self-locking and T/R-SW mechanisms were introduced. The T/R-SW mechanism switches between translational motion to generate self-adaptive finger motion and rotational motion to unlock the self-locking mechanism to return the finger to its original position. A few studies have proposed mechanism designs that can switch motions according to the load [19][20][21]. However, no mechanism has been developed that drives both finger motion and locking using a single actuator.

The contribution of this study is the development of a novel robotic hand that can achieve self-adaptive grasping and a high payload (>20 kg) using only a single low-power motor (6.7 W) with a new motion switching system.

## II. FUNCTIONAL REQUIREMENTS AND KEY MECHANISMS

### A. Functional requirements

The functional requirements of the proposed robotic hand are as follows: 1) payload >20 kg, 2) self-adaptive grasping function, 3) use of a single motor, and 4) lightweight design (<1.0 kg including the motor).

### B. Overview of the developed robotic hand

Figs. 1(a) and 2 show the three-dimensional computer-aided design (3D-CAD) model of the developed robotic hand and its key mechanisms. The three fingers are arranged oppositely to enhance contact area for grasping large/long and heavy objects. Their characteristics and functions are as follows.

***T/R-SW mechanism***: The T/R-SW mechanism switches motion using a single motor in response to an external load. The T/R-SW mechanism generates translational motion of the drive shaft (purple part in Fig. 2(a)) via motor rotation when a small load is applied to the tip of the drive shaft. If a large load is applied and the translational motion is disturbed, the output motion switches from translation to rotation.

***Self-adaptive finger with the self-locking mechanism***: The finger is composed of multiple links and a flexible shaft (red and gray parts in Fig. 2(b)). The finger motion is generated by translating the flexible shaft. When the flexible shaft translates in the direction of the fingertip, bending motion of the finger is induced. The finger bends to conform to the shape of the object

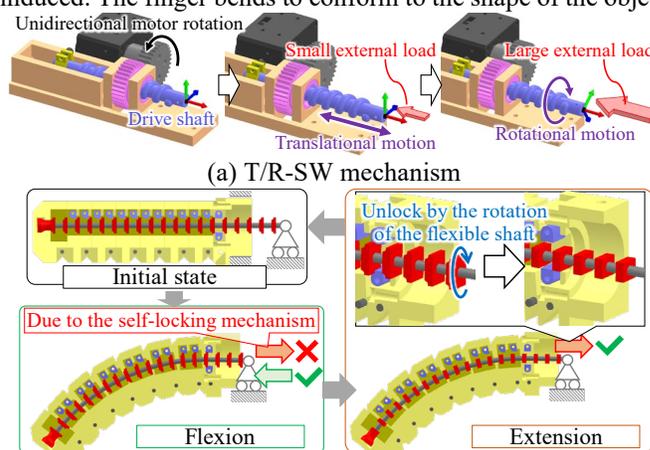

(a) T/R-SW mechanism

(b) Self-adaptive finger with self-locking mechanism

**Fig. 2.** 3D-CAD model of the key mechanisms

owing to its structure, which comprises many links and the deformability of the flexible shaft. The flexible shaft can move in the direction of the finger bending, whereas the self-locking mechanism of the finger prevents motion in the opposite direction. Thus, a high resistance to disturbances to the finger is achieved. The unlocking is accomplished by axially rotating the flexible shaft.

***Synergic effect through integration of the key mechanisms***: The tip of the drive shaft of the T/R-SW mechanism is connected to the flexible shaft of the finger system. Flexion of the finger and self-locking of the bent finger posture are achieved by rotating the motor installed in the T/R-SW mechanism in the extension direction of the tip of the slide axis of the T/R-SW mechanism. The self-locking mechanism can be unlocked for finger extension by rotating the motor in opposite directions. The translational motion of the flexible shaft, i.e., the drive shaft, is prevented by the self-locking function of the finger, and the opposite rotation of the motor causes the rotational motion of the T/R-SW mechanism, which unlocks the self-locking mechanism. Thus, the developed robotic hand employs the T/R-SW mechanism to realize a large payload with a self-locking mechanism and self-adaptability with a flexible shaft and multiple links.

### C. T/R-SW mechanism

#### a) Structure and statics

This section describes the conditions required to achieve the desired behavior of the T/R-SW mechanism, which is self-motion switching. Fig. 3 illustrates the structure mechanism. The T/R-SW mechanism comprises a drive shaft (purple part in Fig. 3) with an outer thread, gear 1 (pink part) with an inner thread, slider (yellow part), motor (black part) with gear 2 (gray part), and base (light orange part). Gear 1 is mounted on the base to allow only axial rotation. The guide rail structure is installed between the slider and base to allow only the linear motion of the slider. The inner thread of gear 1 and outer thread of the drive shaft are engaged. The slider has a C-shaped part with slit width $d_{slit}$. The inner diameter of the C-shaped part is designed to be slightly larger than the outer diameter of the drive shaft. The drive shaft passes through the C-shaped part. The screw is tightened to deform the C-shaped part and clamp the drive shaft until the slit closes completely. Thereafter, a preload is applied to the drive shaft (Fig. 3(a)). The initial width of the slit $d_{slit}$, is the design parameter used to determine the magnitude of the preload. The full closure of the slit ($d_{slit} \rightarrow 0$) for the preloading reduces the difference in the preload value owing to preload resetting and ensures reproducibility of the preload setting. A large $d_{slit}$ value indicates a large preload. The preload causes the frictional torque, $\tau_{pre}$ to be applied to the drive shaft to prevent the shaft from rotating around its longitudinal axis. If the motor is rotated, a thrust force, $f_{sh}$ and rotational torque, $\tau_{sh}$, are applied to the drive shaft through the threads (Fig. 3(b)). If an external load, $f_{ex}$, is applied to the tip of the drive shaft in the axial direction, $f_{sh}$ is generated that counteracts $f_{ex}$, and the corresponding $\tau_{sh}$ is also generated. If $f_{ex}$ is small, the generated $\tau_{sh}$ is counteracted by $\tau_{pre}$; thus, a translational motion of the drive shaft is generated while a rotational motion of the drive shaft is constrained. If $f_{ex}$ is considerably large that the correspondingly generated $\tau_{sh}$ exceeds the maximum $\tau_{pre}$ (maximum static frictional torque owing to the preload), the slippage between the slider and drive shaft occurs, resulting in rotational motion of the drive shaft. Accordingly, self-motion switching can be realized.

The frictional torque applied to the drive shaft by the preload, $\tau_{pre}$, is an essential parameter for realizing self-motion switching. Here, the conditions of the design parameters and $\tau_{pre}$ to achieve switching are described. Fig. 4 shows the mechanical relationship when the input torque, $\tau_{in}$, is applied to gear 1 from the motor through gear 2. The coordinate frame, $\Sigma_A$, is set as shown in Figs. 3 and 4. Let $r_{g1}$ and $r_{g2}$ be the pitch radii of gears 1 and 2, respectively, and let $\tau_m$ be the torque from the motor; then, $\tau_{in}$ can be expressed as

$$\tau_{in} = r_{g1}\tau_m/r_{g2}. \qquad (1)$$

From Newton's third law, $\tau_{in}$ is equal to the torque applied to gear 1 from the drive shaft $\tau_{sh}$:

$$\tau_{in} = \tau_{sh}. \qquad (2)$$

A small contact surface area between the inner thread of gear 1 and the outer thread of the drive shaft is considered. The normal force $df_n$, and frictional/tangential force $df_{fri}$, are applied to the contact surface of the thread. Let $f_n$ and $f_{fri}$ be the normal and tangential forces applied to the entire contact surface of the thread, respectively. These can be expressed as follows:

$$f_n = \int_{s1} df_n, f_{fri} = \int_{s1} df_{fri} \qquad (3)$$

where $s1$ denotes the contact surface area of the thread. Thereafter, the thrust force applied to the drive shaft from gear 1, $f_{sh}$, and $\tau_{sh}$ are expressed as

$$f_{sh} = f_n \cos\theta_{th} - f_{fri} \sin\theta_{th}, \qquad (4)$$
$$\tau_{sh} = r_{g1}(f_n \sin\theta_{th} + f_{fri} \cos\theta_{th}). \qquad (5)$$

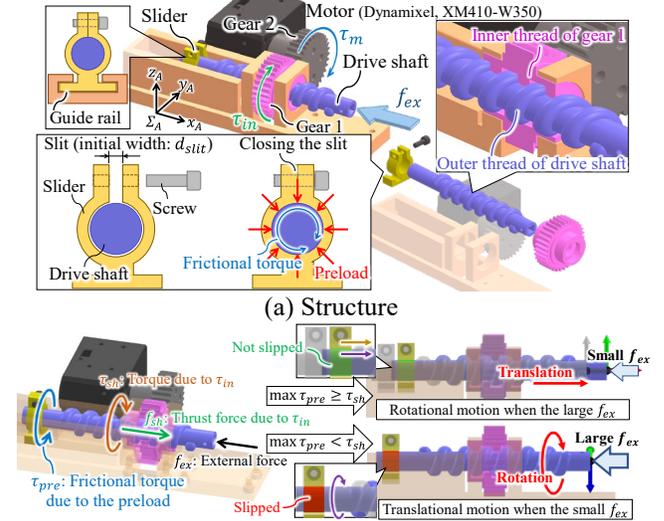

(a) Structure

(b) Typical motions

**Fig. 3.** T/R-SW mechanism

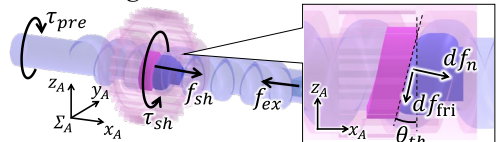

**Fig. 4.** Mechanical relationship

where $\theta_{th}$ is the lead angle of the threads in gear 1 and the drive shaft. If an external force $f_{ex}$, is applied to the tip of the drive shaft, the force and moment balances in the drive shaft are

$$f_{sh} + f_{ex} = 0 \tag{6}$$
$$\tau_{sh} + \tau_{pre} = 0 \tag{7}$$

Considering (2) and (6) and the direction of $f_{sh}$, the following relationship is satisfied in the screw-driving mechanism:

$$\tau_{in} = \tau_{sh} = r_{g1} \tan\theta_{th} f_{sh} = -r_{g1} \tan\theta_{th} f_{ex} \tag{8}$$

To switch from translational to rotational motion of the drive shaft according to the external force ($f_{ex}$), the following conditions should be satisfied: 1) no slippage occurs between the slider and drive shaft, even when the maximum frictional force that can be generated is exerted on the contact surface of the threads, and 2) when the external load is large, slippage occurs between the slider and drive shaft. The first condition is

$$f_{sh}|_{f_{\text{fri}} = \mu_{st} f_n} > |f_{ex}| \tag{9}$$
$$|\tau_{pre}^{st.max}| > \tau_{sh}|_{f_{\text{fri}} = \mu_{st} f_n} \tag{10}$$

where $\mu_{st}$ is the maximum static frictional coefficient at the contact surface of the threads and $\tau_{pre}^{st.max}$ is the maximum static frictional torque at the slider owing to the preload. Here, the kinetic frictional coefficient is assumed to be lower than $\mu_{st}$. Under conditions of (9) and (10), the drive shaft translates in the $x_A$ direction with slippage on the threaded surface, similar to the screw-driving mechanism. If we assume that $f_{ex}$ is sufficiently small, then based on (2), (5), (6), and (8), (9) and (10) can be expressed as

$$\tan\theta_{th} < 1/\mu_{st}, \tag{11}$$
$$|\tau_{pre}^{st.max}| > r_{g1} \tan\theta_{th} |f_{ex}|. \tag{12}$$

(12) can be satisfied by setting the preload value such that $\tau_{pre}^{st.max}$ is large. Hence, by setting $\theta_{th}$, $\mu_{st}$ and the preload value to satisfy (11) and (12), the translational motion is available under a small external load. Similarly, from (2), (7), and (8), the relationship between $\tau_{pre}$ and the external load ($f_{ex}$) can be expressed as follows:

$$\tau_{pre} = r_{g1} \tan\theta_{th} f_{ex} \tag{13}$$

From (13), the larger the external load ($f_{ex}$), the larger the frictional torque ($\tau_{pre}$). When $f_{ex}$ reaches a value that maximizes $\tau_{pre}$ to $\tau_{pre}^{st.max}$, slippage occurs between the drive shaft and slider. In this case, (13) becomes:

$$\tau_{pre}^{st.max} = r_{g1} \tan\theta_{th} f_{ex}. \tag{14}$$

(14) indicates that $\tau_{pre}^{st.max}$ determines the magnitude of $f_{ex}$ that causes slippage. Letting the maximum generable motor torque be $\tau_m^{max}$, from (1) and (14), the second condition is

$$|\tau_{pre}^{st.max}| = r_{g1} \tan\theta_{th} |f_{ex}| \le \tau_{in}|_{\tau_m = \tau_m^{max}} = \frac{r_{g1}}{r_{g2}} \tau_m^{max} \tag{15}$$

(15) indicates that $\tau_{pre}^{st.max}$ should be within the range where the slider slides owing to the motor torque being below its maximum value. By setting the preload value within the range given by (15), motion switching from translation to rotation can be achieved using a single motor.

*b) Validation*

To evaluate the T/R-SW mechanism, an experimental validation was conducted. Fig. 5(a) shows the experimental setup. The O-shaped adapter was assembled with the drive shaft via a bearing so that it could rotate freely. The other side of the O-shaped adapter was fixed to a force gauge. In this setting, the thrust/translational force was applied only to the force gauge. The tip of the drive shaft was equipped with a marker, and a camera was used to monitor the posture of the marker. Using this system, the external force, $f_{ex}$, was measured by the force gauge, the rotational angle, $\theta_{sh}$, of the drive shaft was measured by the camera image using image processing, the rotational angle, $\theta_m$, of the motor was measured by the motor encoder, and the motor torque, $\tau_m$, was measured based on the motor current. To investigate the influence of the preload value, C-shaped sliders with slit widths ($d_{slit}$) of 0, 1, 1.5, 2, and 3 mm were prepared (see also Fig. 3(a)). As described in Section II.C.a, a large $d_{slit}$ value indicates a large preload. The slider with $d_{slit} = 0$ mm applied no preload because the inner diameter of the C-shaped part was designed to be slightly larger than the outer diameter of the drive shaft (Fig. 5(e)). In the experiments, $f_{ex}$, $\theta_{sh}$, $\theta_m$, and $\tau_m$ were measured while the motor was rotated at a constant angular velocity. The results are shown in Figs. 5(b-f). Self-motion switching was achieved for $d_{slit}$ =1, 1.5, and 2 mm. Figs. 5(b-d) show these cases. As $\theta_m$ increased, initially, $f_{ex}$ and $\tau_m$ increased, while $\theta_{sh}$ remained close to zero. Specifically, only the translational motion occurred. When $f_{ex}$ reached its peak value, rotational motion occurred. Correspondingly, $f_{ex}$ and $\tau_m$ decreased. After this moment, $f_{ex}$ and $\tau_m$ became constant, and $\theta_{sh}$ increased proportionally to $\theta_m$. This result shows that translational motion was stopped and rotational motion was generated. Thus, self-motion switching from translation to rotation was realized according to the external load. In addition, the peak value of $f_{ex}$ changed according to the preload value. This demonstrated the validity of (14): the magnitude of $f_{ex}$, which is the threshold for motion switching, can be controlled by $\tau_{pre}^{st.max}$, which is determined by the preload value. When no preload was applied by setting $d_{slit} = 0$, only rotational motion and not translational motion was generated (Fig. 5(e)). This is because (12) was not satisfied as $\tau_{pre}^{st.max}$ was extremely small. By contrast, switching from translation to rotation did not occur when $d_{slit} = 3$ mm (Fig. 5(f)). In this case, (15) was not satisfied because of the extremely large $\tau_{pre}^{st.max}$. These results demonstrate the validity of the T/R-SW mechanism analysis, and show that the desired behavior can be achieved by setting the parameters based on the analysis.

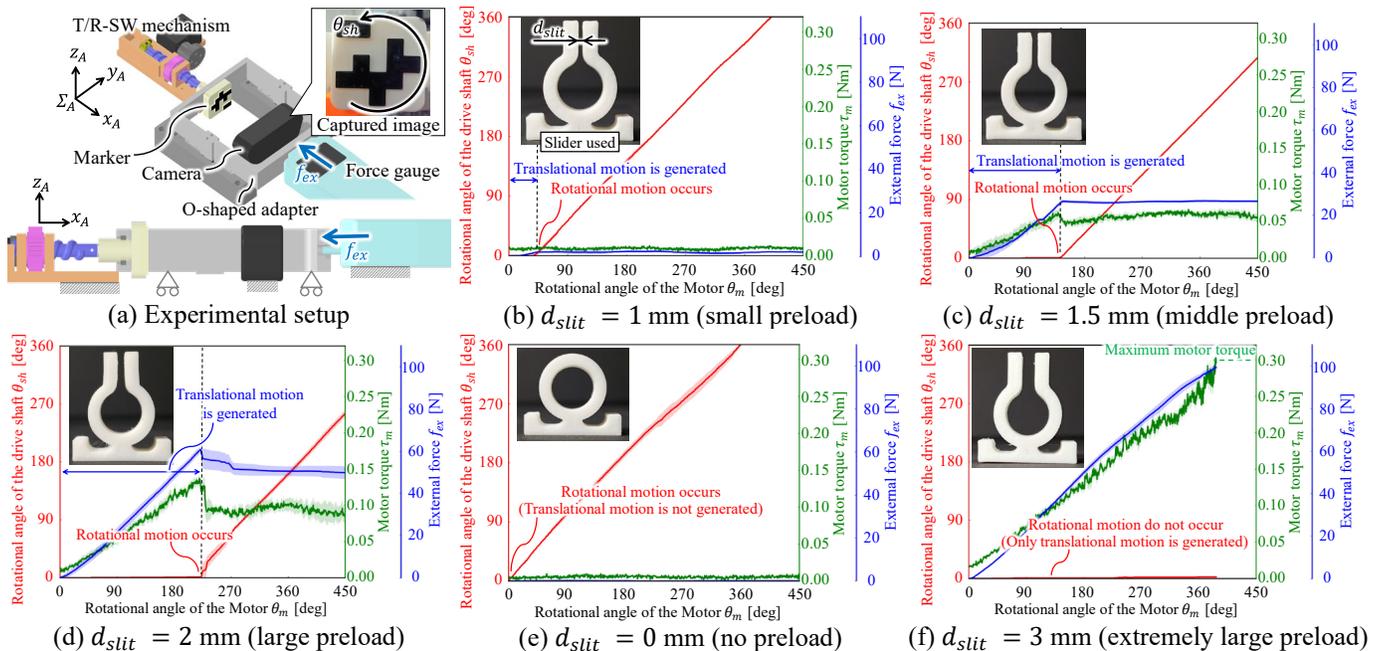

**Fig. 5.** (a) Experimental setup and (b-f) results of the evaluation of the T/R-SW mechanism.

### D. Finger with the self-lock mechanism

This section describes the structure of the developed finger with a self-locking mechanism and the motion analysis to determine the design parameters.

#### a) Structure

Fig. 6 shows the 3D-CAD model of the developed finger. For the explanation, the coordinate frame, $\Sigma_B$, is set as shown in Fig. 6(a). Here, the structure is described by assuming that the finger is in a nominal state when it is extended. The finger comprises a multilinked body and a flexible shaft with multiple protrusions. As shown in Fig. 6(b), the links composing the finger body (FB link: yellow part in Fig. 6) have a hollow structure, and are connected via joint pins (gray part) and torsion spring 1 (brown part). The finger motion behavior is determined by the spring coefficients of the torsion springs. The details of this process are described in the next section. Pawls (purple part) are installed inside the FB links via torsion spring 2. The FB link incorporates a stopper (yellow) to partially constrain the rotation of the pawls. The posture of the pawl is maintained such that the long side of the pawl is in the $y_B$ direction because of the restoring torque provided by torsion spring 2 and stopper (if nothing is in contact with the pawl). This is referred to as the nominal posture of the pawl. Because of the stopper, the pawl in the nominal posture can only rotate around the $+z_B$ axis (Fig. 6(b)). A flexible shaft is installed such that it passes through the finger body, and is connected to the most distal FB link with the free rotational joint. The shaft is comprised of a metal wire rope (diameter: 2.5 mm), and its protrusions are made of resin (IMT, Zeromer K+) (Fig. 6(c)). The flexibility in the bending direction and high torsional stiffness to rotate the shaft are required as the deformable elements for the flexible shaft. In addition, the shaft should be sufficiently stiff to prevent buckling owing to bending. Based on these requirements, the metal wire was used in this study. In the fabrication of the flexible shaft, the metal rope was placed in the mold and the resin was injected. The metal rope was demolded after curing the resin, and the burr of the cured resin was removed. The flexible shaft only allows bending owing to its high torsional stiffness. The state of contact between the protrusions and pawls changes according to the rotational angle of the flexible shaft in the axial direction because the protrusion is rectangular as shown in Fig. 6(d). The locking mechanism uses variable contact states to achieve self-locking and unlocking. When the long side of the protrusion is parallel to the $y_B$-axis as shown in Fig. 6(d), contact between the protrusion and pawl occurs; this state is referred to as the self-locking mode. In contrast, when the short side is parallel to the $y_B$-axis, the protrusion cannot make contact with the pawl; this state is referred to as the unlocking mode. In the initial state, the locking mechanism was set to self-locking mode (Fig. 6(a)). The subsequent behavior of the finger is illustrated in Fig. 7. The translational motion of the flexible shaft in the $+x_B$ direction increases the total length of the area where the flexible shaft is inserted while maintaining the length of the area connected by the joint pins. This difference results in finger flexion. The translational movement of the flexible shaft is not prevented by the pawls, as shown in Fig. 7. The protrusions of the flexible shaft rotate the pawls that return to their nominal positions after the protrusions pass through them. In contrast, translation in the $-x_B$ direction, i.e., the direction of finger extension, is prevented by contact between the protrusions and pawls because the pawls cannot rotate around the $-z_B$ axis because of the stoppers. In this structure, the finger posture is firmly maintained even when a large external load is applied, and a self-locking mechanism is achieved. By rolling the flexible shaft around its axial direction to switch from the self-locking mode to the unlocking mode, the contacts between the protrusions and pawls are released. Thus, unlocking is achieved. The extension of the finger can be achieved by translating the flexible shaft in the $-x_B$ direction after unlocking. As the pawls were installed discretely, there are finger postures in which the locking mechanism does not work, although the rigidity of the flexible shaft allows the finger posture to be maintained. If a large external load is applied to the finger that is not locked, the finger posture would be displaced to the position where the locking mechanism is activated. To reduce the displacement, two pawls were installed per FB link, as shown in Fig. 6(d), increasing the possibility that any of the pawls will contact the protrusions.

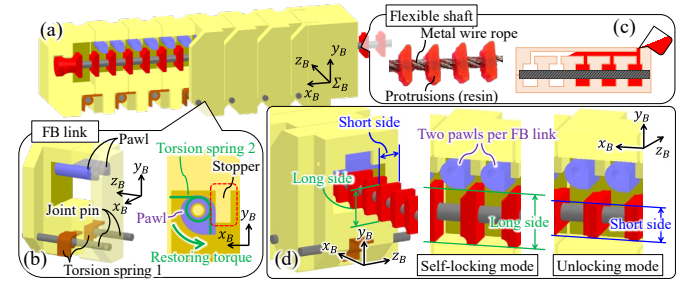

**Fig. 6.** Structure of the finger with the self-locking mechanism. (a) Entire design. (b) FB link. (c) Structure and fabrication method of the flexible shaft. (d) States of the lock mechanism.

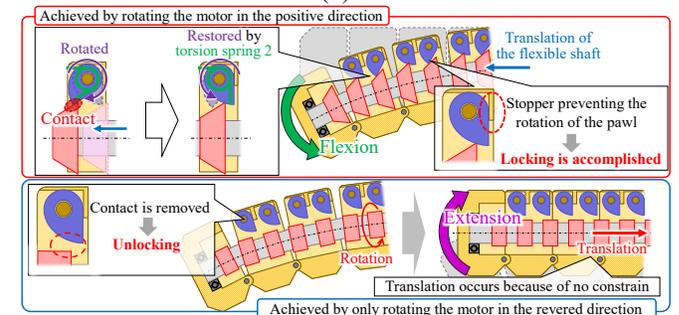

**Fig. 7.** Behavior of the finger with the self-locking function

*b) Analysis*

This section presents the motion analysis of finger bending. The coefficients of torsion spring 1 installed on the joints of the FB links, which affect finger behavior, were determined based on this analysis. The model of the $n$-linked finger used in the analysis is illustrated in Fig. 8. The relationship between the force applied to push the flexible shaft into the finger $f_{tr}$, and the resultant finger posture was examined. In this study, a planar analysis was performed. As shown in Fig. 8, the reference coordinate frame for this analysis, $\Sigma_C$, was set as follows: 1) the origin was placed at the most proximal joint pin of the FB links; 2) the positive direction of the $x_C$ axis corresponded to the longitudinal direction from the origin to the fingertip in the initial state; and 3) the positive direction of the $y_C$ axis was directed from the origin to the back side of the 1st FB link. Let $\boldsymbol{p}_{L_i} \in \mathbb{R}^2$ be the position of the joint pin between the $i$th and $(i-1)$th FB links ($i \in \{1, 2, \cdots, n\}$). Note that $\boldsymbol{p}_{L_1}$ corresponds to the origin of $\Sigma_C$. If the flexible shaft is translated by the force $\boldsymbol{f}_{tr} = [f_{tr}, 0]^\mathrm{T}$, the finger is bent. In the $i$th FB link, the following forces and torques are applied: the forces and torques from the two adjacent FB links and the force from the flexible shaft. Let $\theta_{L_i}$ be the angular displacement between the $i$th and $(i-1)$th FB links, and $\boldsymbol{f}_{L_i} \in \mathbb{R}^2$ and $m_{L_i} \in \mathbb{R}$ be the force and torque applied to the connecting joint at $\boldsymbol{p}_{L_i}$, respectively. This section aims to derive the relationship between the posture of the finger, i.e., $\theta_{L_i}$, and the force required to bend the finger ($f_{tr}$), which is controlled by the motor torque through the T/R-SW mechanism. This is designed such that the FB links connected via the joint pins can rotate freely around the joint pins, although their rotation is constrained by torsion spring 1 and the flexible shaft. Let $m_{L_i}$ be the restoring torque by the torsion spring and flexible shaft, and $k_{L_i}$ be the composite torsional spring coefficients of torsion spring 1 and flexible shaft. Subsequently, $m_{L_i}$ is expressed as

$$m_{L_i} = k_{L_i} \theta_{L_i}. \tag{16}$$

Letting $k_{sp_i}$ be the coefficient of torsion spring 1 installed in the joint, $\boldsymbol{p}_{L_i}$, and $k_{FS}$ be the flexural stiffness of the flexible shaft, $k_{L_i}$ can be expressed as follows:

$$k_{L_i} = k_{sp_i} + k_{FS} \tag{17}$$

In the $i$th FB link, the coordinate frame $\Sigma_{L_i}$ is set such that 1) the origin is located at $\boldsymbol{p}_{L_i}$; 2) the $x_{L_i}$ axis is directed from $\boldsymbol{p}_{L_i}$ to $\boldsymbol{p}_{L_{i+1}}$; and 3) the positive direction of the $y_{L_i}$ axis is directed from $\boldsymbol{p}_{L_i}$ to the back side of the $i$th FB link (Fig. 8). The flexible shaft is assumed to be parallel to the $x_{L_i}$ axis for each FB link. In addition, the force acting between the $i$th FB link and flexible shaft, $\boldsymbol{f}_{FS_i} \in \mathbb{R}^2$, is assumed to be generated owing to the friction between the flexible shaft and FB link because the contacting force between the protrusion and pawl is sufficiently small owing to the freely rotatable pawls. Denoting the vector with respect to $\Sigma_{L_i}$ by the upper-left superscript "$i$," from the above assumptions, ${}^i\boldsymbol{f}_{FS_i}$ is expressed as

$${}^i\boldsymbol{f}_{FS_i} = [f_{FS_i} \quad 0]^\mathrm{T}, (f_{FS_i} > 0) \tag{18}$$

Thereafter, the force and moment balances of the $i$th FB link are expressed as

$${}^i\boldsymbol{f}_{L_i} - {}^i\boldsymbol{R}_{i+1}{}^{i+1}\boldsymbol{f}_{L_{i+1}} + [f_{FS_i} \quad 0]^\mathrm{T} = \boldsymbol{0} \tag{19}$$

$$m_{L_i} - m_{L_{i+1}} + d_L f_{FS_i} \\ - \left( {}^i\boldsymbol{p}_{L_{i+1}} - {}^i\boldsymbol{p}_{L_i} \right) \otimes {}^i\boldsymbol{R}_{i+1}{}^{i+1}\boldsymbol{f}_{L_{i+1}} = 0 \tag{20}$$

where $d_L$ is the distance between the rotational joint ($\boldsymbol{p}_{L_i}$) and flexible shaft. ${}^i\boldsymbol{R}_{i+1}$ is the rotational matrix expressing the posture of $\Sigma_{L_{i+1}}$ with respect to $\Sigma_{L_i}$:

$${}^i\boldsymbol{R}_{i+1} = \begin{bmatrix} \cos\phi_i & -\sin\phi_i \\ \sin\phi_i & \cos\phi_i \end{bmatrix}, \text{where } \phi_i = \sum_{i=i+1}^{n} \theta_{L_i} \tag{21}$$

$\otimes$ in (20) represents the following cross-product operation:

$$[x_1 \quad y_1]^\mathrm{T} \otimes [x_2 \quad y_2]^\mathrm{T} = x_1 y_2 - x_2 y_1. \tag{22}$$

If $l_L$ is the length between the connecting joints ($l_L$ is the same for each FB link), then $\left( {}^i\boldsymbol{p}_{L_{i+1}} - {}^i\boldsymbol{p}_{L_i} \right)$ in (20) is

$$\left( {}^i\boldsymbol{p}_{L_{i+1}} - {}^i\boldsymbol{p}_{L_i} \right) = [l_L \quad 0]^\mathrm{T}. \tag{23}$$

In the $n$th FB link, which connects to the other link on only one side, the force and moment balances are expressed as

$${}^n\boldsymbol{f}_{L_n} - [f_{FS_n} \quad 0]^\mathrm{T} = \boldsymbol{0} \tag{24}$$

$$m_{L_n} + d_L f_{FS_n} = 0 \tag{25}$$

From (16) and (25), the posture of the $n$th FB link, $\theta_{L_n}$, is

$$\theta_{L_n} = d_L f_{FS_n} / k_{L_n} \tag{26}$$

$k_{sp_i}$, $k_{FS}$, $d_L$, and $l_L$ are design parameters. Hence, if ${}^i\boldsymbol{f}_{FS_i} = [f_{FS_i}, 0]^\mathrm{T}$ is obtained, the posture of all FB links, $\theta_{L_i}$, can be derived using (19) and (20). As described above, $f_{FS_i}$ corresponds to the frictional force between the flexible shaft and FB link, and is applied in the axial direction of the flexible shaft. Thus, the following equation is satisfied.

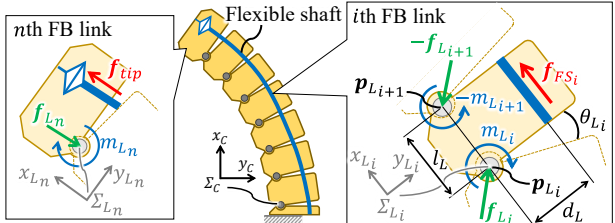

**Fig. 8.** Analysis model

$$\int_{s2} \boldsymbol{f}_{FSi} = \sum_{i=1}^{n} f_{FSi} = f_{tr} \quad (27)$$

where $s2$ denotes the flexible shaft path. $f_{FSi}$ can be derived by minimizing the elastic energy of the flexible shaft and torsion springs 1 under the constraints of (19), (20), and (27) as follows:

$$f_{FSi} = \underset{f_{FSi}}{\operatorname{argmin}} \frac{1}{2} \sum_{i=1}^{n} k_{Li} \theta_{Li}^{2} \quad (28)$$

*Subject to* (19), (20), and (27) for given $f_{tr}$

The postures of the FB links, $\theta_{Li}$, are obtained through the deviation process in (28).

Next, the methodology for determining the coefficients of torsion spring 1 to achieve the desired finger motion behavior is described. The parameters were set as follows: $n = 7$, $d_L = 13$ mm, and $l_L = 12$ mm. First, the flexural stiffness of the flexible shaft was investigated. Assuming that the flexible shaft acts as a torsion spring, as shown in (17), the stiffness of the flexible shaft was determined experimentally. The experimental setup is shown in Fig. 9. In the experiment, a force gauge (IMADA, ZTS-500N) was used to push the flexible shaft and measure the pushing force $f_{tr}$. To investigate the effect of the flexible shaft, torsion spring 1 was not installed at the connecting joints. The positions of the joint pins $\boldsymbol{p}_{Li}$, were measured using image processing when a pushing force was applied. The stiffness of the flexible shaft was calculated:

$$k_{FS} = \underset{k_{FS}}{\operatorname{argmin}} \sum_{i=1}^{n} \left\| \boldsymbol{p}_{Li}^{th}(k_{FS}) - \boldsymbol{p}_{Li}^{act} \right\| \quad (29)$$

where $\boldsymbol{p}_{L}^{th}$ is $\boldsymbol{p}_{Li}$ theoretically calculated using $\theta_{Li}$ ( $i \in \{1, \cdots, n\}$ ) from the deviation process of (28), and $\boldsymbol{p}_{Li}^{act}$ is the measured $\boldsymbol{p}_{Li}$. The measurements were conducted under the condition where the values of $f_{tr}$ were set to 3.2 N, 5.1 N, and 6.3 N. Fig. 10 shows the results of the calculated $k_{FS}$, and the calculated and measured postures for each $f_{tr}$. The $k_{FS}$ values in each condition were close to each other, allowing accurate posture estimation (Fig. 10). To determine the coefficients of torsion spring 1 ( $k_{Li}$ ), $k_{FS}$ was set as the mean of three calculated values, i.e., $k_{FS}$ =4.5 Nmm/deg. With this value, $k_{Li}$ was determined such that all the joints of the finger were bent uniformly, as shown in Fig. 1(c).

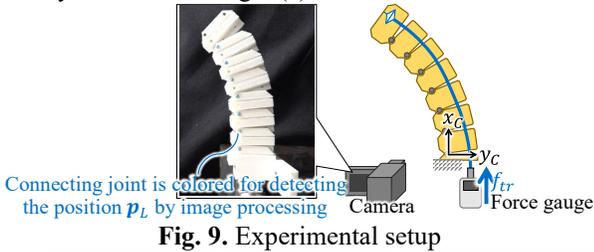

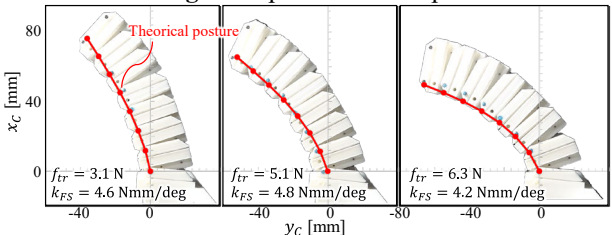

**Fig. 9.** Experimental setup

**Fig. 10.** Identification of $k_{FS}$. Results of the observed and theoretical finger postures

## III. EVALUATION OF The Finger with Key Mechanisms

This section presents the evaluation results for the developed finger. Based on the experiments shown in Fig. 5, $d_{slit}$ was set as 2 mm for the T/R-SW mechanism.

### A. Adaptability

The finger's adaptability was evaluated by the experiments in which the finger was wrapped around target objects of different diameters (20, 40, and 60 mm). In the experiments, the motor rotated until the motor torque ($\tau_m$) reached the threshold $\tau_{th}$, which was within the range where rotational motion was not activated. $\tau_{th}$ was set to 0.1 Nm in this experiment. After the motor was stopped, the finger posture was observed. The results are presented in Fig. 11(a). The fingers achieved self-adaptive motion for cylinders with diameters of 20 and 40 mm. However, adaptation was insufficient for the 60-mm-diameter object because the fingertip contacted the object before adaptation was completed. The experiments were also conducted when $k_{Li}$ was set to bend the finger from its proximal end; the results are shown in Fig. 11(b). This setting enabled the high adaptation. The results demonstrated that the finger provided the self-adaptive function, and adaptability could be varied by parameter settings.

### B. Resistivity

The experimental setup used to evaluate resistivity is shown in Fig. 12(a). After the finger was bent by rotating the motor until $\tau_m = \tau_{th}$, an external force was applied to the finger through the object. When the external force reached 104.2 N, the FB link of the finger broke. This result demonstrates that the finger can resist a disturbance of more than 100 N. Fig. 12(b) shows the broken FB link. Damage occurred at the connecting point between the FB link and joint pin. The stress tends to concentrate at this point because of its shape. The resistivity can be improved by increasing the rigidity of this part.

## IV. GRASPING TEST

This section describes the results of the grasping tests. The developed robotic hand was attached to the automatic positioning stage and the manipulator (UR5 and Yaskawa GP7) grasped the objects placed on a table. Various objects, including heavy (a 22.5-kg-dumbbell and 14-kg-air compressor), large (a stepladder and box), and deformable (a bag containing dried noodles) objects, were successfully grasped, as shown in Figs. 1(a) and 13. The large box was lifted by catching the overhanging edge with only two fingers, thereby utilizing its high resistance to disturbing forces (Fig. 13(c)). The self-adaptability allowed the deformable bag to be grasped (Fig. 13(d)). A grasping test of a 30-kg-dumbbell was also conducted, which resulted in failure owing to breakage of the finger, despite the fact that the theoretical maximum payload was calculated to be over 30 kg in Section III.B. The results presented in Section III.B demonstrated that the single finger could resist a force of 10 kgf. This failure could be because the external force owing to the grasping of the dumbbell was not applied to the three fingers uniformly; therefore, a large force of over 10 kgf was applied to one finger. The broken part of the FB link was the same as that shown in Fig. 12(b). We replaced the FB links made from PLA with those made from high-stiffness resin (Form 3, tough2000) and repeated the grasping test on a 30-kg-dumbbell. The grasping was successful at this time (see the video clip). These results demonstrated the grasping ability of the developed robotic hand.

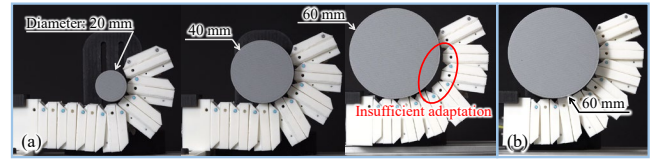

**Fig. 11.** Results of the self-adaptability test. (a) Uniform finger bending. (b) Finger bent from the proximal side

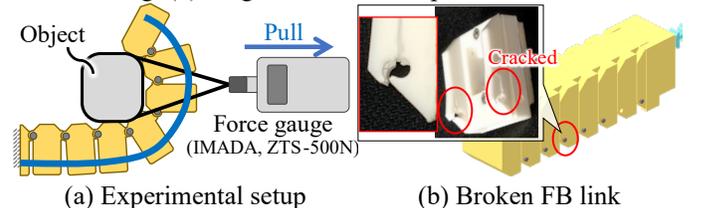

**Fig. 12.** Experimental setup and evaluation results for the disturbance resistivity of the finger

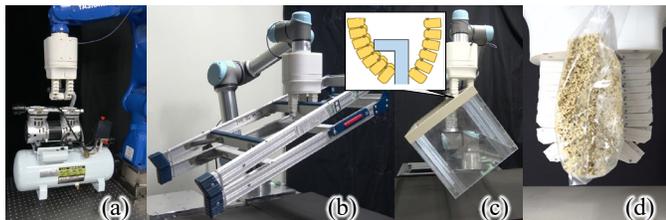

**Fig. 13.** Representative results of the grasping tests. (a) Heavy air compressor (14 kg). (b) Large stepladder. (c) Large box. (d) Deformable bag

## V. CONCLUSION

This paper proposed a novel 1-DOF lightweight (0.8 kg) robotic hand that could achieve a high payload (over 20 kg) and self-adaptive grasping. The robotic hand has two key mechanisms: the T/R-SW mechanism and self-adaptive finger with a self-locking mechanism. The T/R-SW mechanism switches the output motion from translation to rotation according to the external load. The proposed finger comprises a flexible shaft and multiple finger links, allowing self-adaptation during finger flexion. The flexible shaft has multiple protrusions and contacts with the pawls installed inside the finger-link body. This contact limits the translational motion of the flexible shaft for finger extension, and a self-locking function can be achieved to obtain high resistance to the disturbance. The locking mechanism is unlocked by rotating the flexible shaft because of the rectangular protrusions. The flexible shaft is connected to the drive shaft of the T/R-SW mechanism. The flexible shaft is translated by rotating the motor of the T/R-SW mechanism forward, thereby bending the finger. The rotation of the motor in the opposite direction causes the rotation of the flexible shaft because the self-locking mechanism restricts its translational motion, and the motion of the T/R-SW mechanism is switched to rotation. Hence, the proposed robotic hand with these key mechanisms can realize self-adaptability and a high payload with only 1-DOF control, i.e., only a single motor. Analyses of the key mechanisms were presented herein. The design parameters, especially the preload value of the T/R-SW mechanism and the spring coefficient of the finger in the mechanisms, were determined to realize the desired behavior based on the analyses. The adaptability to the object shape and resistance to the external force of the developed finger with the key mechanisms were experimentally validated. A three-fingered robotic hand was also evaluated, and the results demonstrated that all design requirements were satisfied. Our future work will focus on developing a control methodology for grasping and manipulating objects by using the developed robotic hand. Optimization of the design parameters, including the size of the FB link, would be also considered in our future work.